\documentclass[10pt,twocolumn,letterpaper]{article}
\usepackage[pagenumbers]{cvpr} 

\definecolor{cvprblue}{rgb}{0.21,0.49,0.74}
\usepackage[pagebackref,breaklinks,colorlinks,allcolors=cvprblue]{hyperref}
\def\confName{CVPR}
\def\confYear{2027}
\title{DynaMimic: A Physics-Informed World Model for Humanoid Motion Tracking}
\author{
Chaoyue Xing$^{1,2}$ \quad
Linyi Huang$^{3}$ \quad
Yiting He$^{2,3}$ \quad
Yixuan Yin$^{2,4}$ \quad
Siqi Hou$^{1}$ \\[0.5ex]
Yi Zhang$^{2}$ \quad
Liyuan Pan$^{4}$ \quad
Miaomiao Liu$^{1}$ \quad
Daqi Liu$^{2}$ \\[1ex]
{\small $^{1}$The Australian National University
\quad $^{2}$AgiBot} \\
{\small $^{3}$The Hong Kong University of Science and Technology (Guangzhou)}  \\
{\small $^{4}$Beijing Institute of Technology}
}
\usepackage{amsmath,amssymb}
\usepackage{graphicx}
\usepackage{caption}
\usepackage{booktabs,microtype}
\newsavebox{\newmotionpdf}
\newsavebox{\firstmotionrow}
\newsavebox{\cleanmotionrow}
\newsavebox{\motionpanel}

\makeatletter
\renewcommand\paragraph{\@startsection{paragraph}{4}{\z@}{4pt}{-1em}{\normalfont\normalsize\bfseries}}
\makeatother

\begin{document}
\raggedbottom
\setlength{\parskip}{0pt}

\twocolumn[{%
  \renewcommand{\twocolumn}[1][]{#1}%
  \maketitle
  \begin{minipage}{\textwidth}
    \captionsetup{type=figure}
    \centering
    \includegraphics[width=\linewidth]{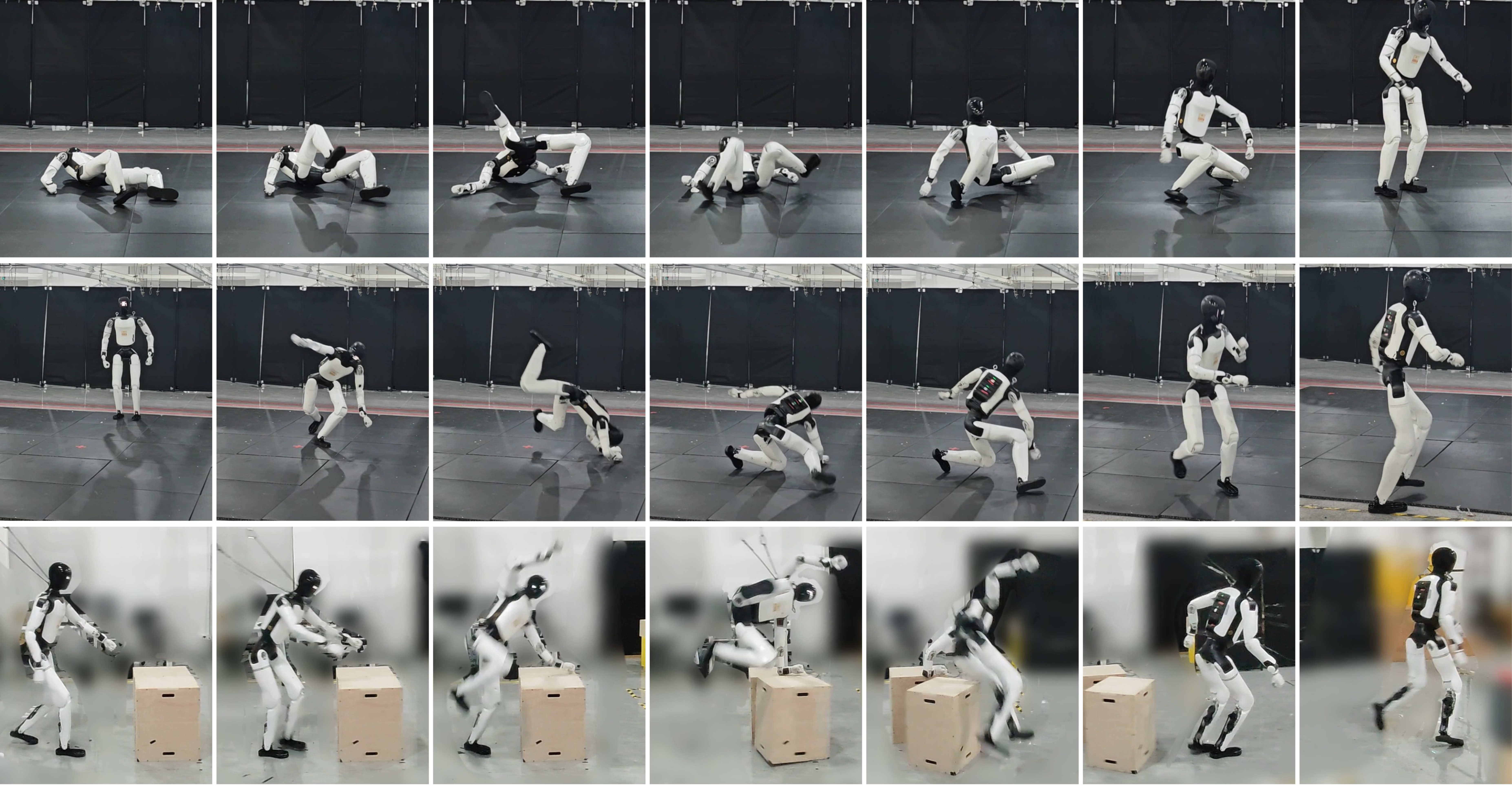}\par
    \caption{\textbf{DynaMimic on an A3 humanoid.}
      Each row shows one motion from left to right.
      The robot rises from the floor (top row),
      performs a cartwheel (middle row),
      and vaults over a box (bottom row).}
    \label{fig:world-model}
  \end{minipage}\par\medskip
}]

\begin{abstract}
Whole-body motion tracking enables humanoid robots to reproduce diverse
skills from human demonstrations. Existing model-free reinforcement
learning methods either distill motion specialists into a single policy
at the cost of performance or train a generalist policy directly using
cluster-scale computation. SuperMimic reduces this computational cost
through a differentiable world model that implicitly learns the physics
shared across motions by predicting the next state from the current
state and action. However, this implicit formulation may not fully
capture the underlying physical processes. Errors around changes of support can accumulate over imagined rollouts used to train the tracking policy.
To address this challenge, we introduce DynaMimic, a differentiable
world model that explicitly incorporates shared physics into state
prediction. Specifically, its network predicts a contact wrench at
each candidate support link and a residual generalized force.
DynaMimic then uses these predictions to compute the next state
through closed-form Lagrangian forward dynamics. 

Across 60 full LAFAN1 motions, DynaMimic reduces mean per-joint position error (MPJPE) by 18.6\% relative to SuperMimic. For terrain tracking, we give both methods the same local height maps. Relative to terrain-conditioned SuperMimic, DynaMimic reduces MPJPE by 14.9\% on A3 Parkour (66.15 to 56.28~mm) and by 10.8\% on GRAIL terrain clips (141.88 to 126.57~mm). On a real A3 humanoid, the learned policies can perform diverse skills, such as floor recovery, cartwheel and a box vault.
\end{abstract}

\vspace{-3mm}
\section{Introduction}
\label{sec:intro}
Whole-body motion tracking allows a humanoid to reproduce diverse skills from human demonstrations~\cite{he2024omnih2o,wang2026omnixtreme}. By following different reference motions, a single tracking policy can enable the humanoid to walk across a room, get up from the floor, or vault over an obstacle. However, training such a policy across a large motion library remains computationally expensive. Existing model-free reinforcement learning methods either distill motion specialists into a single policy at the cost of performance~\cite{liao2026beyondmimic} or train a generalist policy directly using cluster-scale computation~\cite{chen2025gmt,luo2026sonic}.

Recently, SuperMimic~\cite{liao2026supermimic} reduces this computational cost through a differentiable world model that implicitly learns the contact effects, underlying dynamics shared across motions by predicting the next state from the current state and action. However, this implicit mapping is difficult to learn. For example, during floor recovery, extending the elbow can raise the torso while the palm is planted, whereas a similar action mainly moves the arm after the hand leaves the ground. Prediction errors around such contact transitions can propagate through imagined rollouts and affect the gradients used to optimize the policy. Although the contact forces change, the body's response follows the same governing equations of motion. Given the current state, actuator forces, and contact forces, the robot's known forward dynamics determine its acceleration. This motivates learning the interaction-dependent forces and using known mechanics to compute their effects across motions.

To address these challenges, we introduce \textbf{DynaMimic}, a differentiable world model that explicitly incorporates this shared physics into each prediction (Figure~\ref{fig:dynamimic-pipeline}). Given the current state and action, our differentiable world model predicts a contact wrench, comprising a force and a moment, at each candidate support link. It also predicts a residual generalized force to account for effects such as unmodeled joint friction. DynaMimic combines these predictions with the actuator forces, computes the robot's acceleration through Lagrangian forward dynamics, to obtain the next pose and velocity in closed form. This formulation grounds next-state predictions in the robot's equations of motion. Besides the commonly used trajectory loss, we directly supervise the contact wrenches prediction at each candidate support link by using contact wrenches already computed by the simulator, requiring no additional manual annotation.

We further extend DynaMimic to handle a more realistic setting human--scene interaction (HSI), where predicting the effect of an action also requires information about the scene. We represent the local scene with a terrain height map centered on the robot and aligned with its heading. The height map is utilised in the policy training for selecting an action and world model for better motion prediction.

Overall, our contributions are twofold. i) We proposed DynaMimic, a novel method with a physically-informed differentiable world model that combines learned contact wrenches with the robot's known forward dynamics in each state update. ii) We extend DynaMimic to human--scene interaction by conditioning both the tracking policy and the world model on local terrain geometry and updating the terrain input throughout imagined rollouts.~We evaluate DynaMimic on benchmark datasets and outperform the State-of-the-art method consistently. We also deploy the learned policy on a physical humanoid to demonstrate the superiority of our proposed method.

\section{Related Work}
\label{sec:relatedwork}

\paragraph{Humanoid motion tracking.}



DeepMimic~\cite{peng2018deepmimic} used proximal policy optimization (PPO) to train simulated characters to track reference motions. PHC~\cite{luo2023perpetual} extended tracking to large motion libraries with a single policy, and PULSE~\cite{luo2024universal} distilled tracking skills into a latent action space for downstream control. MaskedMimic~\cite{tessler2024maskedmimic} enables control from partial references, including sparse joint targets and keyframes. On physical humanoids, H2O and OmniH2O use human motion for whole-body teleoperation~\cite{helearning,he2024omnih2o}, while HOVER~\cite{he2025hover} supports multiple control interfaces with one policy. ASAP~\cite{he2025asap} and BeyondMimic~\cite{liao2026beyondmimic} further demonstrate agile motion tracking on hardware.

Off-policy methods reuse simulator transitions through experience replay. FlashSAC~\cite{KimD-RSS-26} applies this approach to high-dimensional robot control, and Seo et al.~\cite{seo2025learning} use off-policy training for humanoid locomotion and motion tracking. Model-based methods also use collected transitions to learn how a robot responds to an action, as in DayDreamer~\cite{pmlr-v205-wu23c} and Robotic World Model~\cite{li2025robotic}. For humanoid tracking, SuperMimic~\cite{liao2026supermimic} fits one differentiable dynamics model across reference motions and updates the policy through short imagined rollouts. Its transition network learns the combined effects of actuation, contact, and body dynamics from state transitions.

\paragraph{Physics-informed world model learning.}

Contact provides information about how a body moves within a scene. In human motion recovery, PROX~\cite{hassan2019resolving} uses body--scene contact to constrain pose, and WHAM~\cite{shin2024wham} uses estimated foot contact to refine the global trajectory and reduce foot sliding. Body--scene distances also describe interactions in motion forecasting and generation~\cite{xing2024scene,yang2026learning}. Other methods model the forces underlying these interactions. Xie et al.~\cite{xie2021physics} infer forces consistent with motion observed in video, PhysPT~\cite{DBLP:conf/cvpr/ZhangKCJ24} uses contact and dynamics constraints to estimate motion and forces, and InterPhys~\cite{xing2026interphys} models contact forces during motion synthesis in dynamic scenes.

Hybrid simulators use learned forces to advance the state of a physical system. NeuralSim~\cite{9560935} embeds learned components in differentiable rigid-body dynamics, and Jiang et al.~\cite{pmlr-v168-jiang22a} combine analytical contact mechanics with data to predict the net contact impulse on a rigid body.

DynaMimic uses a hybrid transition within the imagined rollouts that train a humanoid tracking policy. At each step it predicts a wrench for each candidate support link and a residual generalized force, then integrates their effects using the robot's known dynamics. Simulator-recorded link wrenches supervise how the load is distributed across the body, while the state trajectory supervises the resulting motion. We optimize the tracking policy by differentiating its return through both the predicted contact wrenches and the rigid-body dynamics. For terrain-conditioned tracking, both the policy and world model receive local height maps that are updated as the predicted robot pose changes along imagined rollouts.

\section{Method}
\label{sec:method}

\paragraph{Overview.}
DynaMimic is a humanoid motion tracking framework that resolves contact ambiguity by learning a physics-informed world model.
Following SuperMimic~\cite{liao2026supermimic}, we improve the policy by differentiating the tracking objective through short rollouts in a learned world model. Our world model moves beyond standard transition models that implicitly memorize state changes by first predicting the contact wrenches and residual forces from the robot's current state and the policy's action. It then uses rigid-body dynamics to compute the next state. We first introduce the training loop shared across reference motions (Sec.~\ref{sec:formulation}). We then describe how DynaMimic computes a state transition and how we fit it to simulator trajectories (Sec.~\ref{sec:dynamics}). Finally, we use the fitted model to optimize the tracking policy (Sec.~\ref{sec:training}).

\begin{figure*}[t]
\centering
\includegraphics[width=\linewidth]{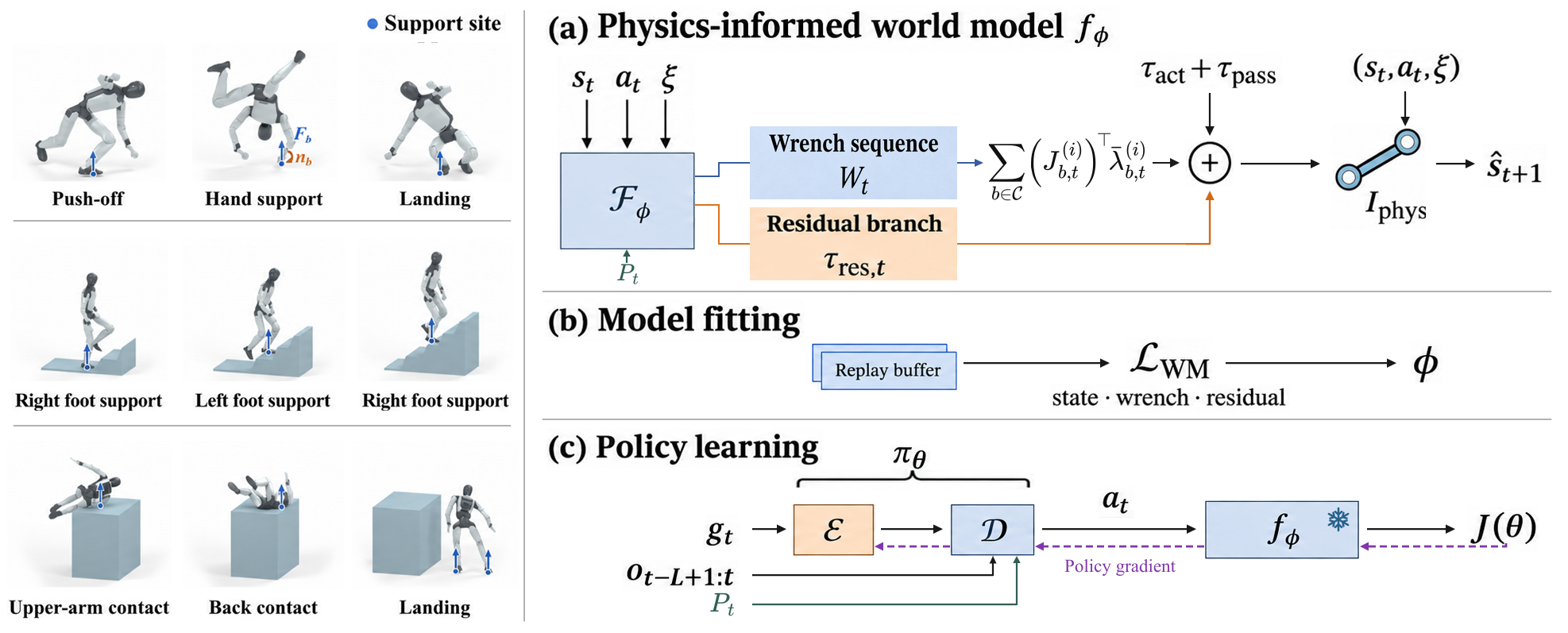}
\caption{\textbf{DynaMimic.} Our framework integrates a differentiable physics-informed world model to learn humanoid motion tracking. (a) The force predictor $\mathcal F_\phi$ predicts contact wrench sequences $W_t$ at candidate support links (blue dots) and a residual generalized force $\tau_{\mathrm{res},t}$. At each substep, the integrator maps each link's mean wrench $\bar\lambda_{b,t}^{(i)}$ to generalized forces, adds the residual, actuator, and passive forces, and advances the state. (b) We fit the world model to simulator states and wrenches from replay and regularize the residual. (c) The reference encoder $\mathcal E$ and action decoder $\mathcal D$ form the policy $\pi_\theta$. Dashed arrows trace the gradient of the imagined return $J(\theta)$ through the frozen world model (snowflake) to the policy. For terrain-conditioned tracking, the policy and world model encode the same local height map $P_t$ with separate CNNs.}
\label{fig:dynamimic-pipeline}
\end{figure*}

\subsection{Training Loop}
\label{sec:formulation}
%
DynaMimic trains a whole-body tracking controller in one alternating loop of simulator data collection, physics-informed world-model fitting, and imagination-based policy optimization (Fig.~\ref{fig:dynamimic-pipeline}).
Training alternates between three steps. First, the current policy acts in simulation, and we store the resulting trajectories in a replay buffer together with the executed actions and measured link wrenches. Second, we fit \(f_\phi\) to short windows from the buffer by rolling it forward under the recorded actions. Third, we fix the world-model parameters and let the policy select its own actions over short imagined rollouts. We update the policy by differentiating the tracking return through these rollouts. The updated policy then collects the next batch of simulator data.

Formally, at control step \(t\), the policy \(\pi_\theta\), with parameters \(\theta\), selects an action \(a_t\in\mathbb R^{29}\) specifying targets for the actuated joints. Its inputs are the last \(L=10\) proprioceptive observations, written \(o_{t-L+1:t}=(o_{t-L+1},\ldots,o_t)\), and a window \(g_t\) of ten reference frames containing joint positions, joint velocities, and relative base orientations. Let \(s_t\) denote the robot state and \(\xi\) the episode's friction coefficient, joint calibration offsets, and torso center-of-mass displacement. On level ground, the policy and the state transition \(f_\phi\) of the world model take the form
\begin{equation}
\begin{aligned}
a_t &= \pi_\theta(o_{t-L+1:t},g_t),\\
\hat s_{t+1} &= f_\phi(s_t,a_t,\xi).
\end{aligned}
\label{eq:method-io}
\end{equation}
Here \(\phi\) denotes the trainable world-model parameters. A hat marks a model prediction, and a superscript \(\star\) marks a simulator target in the losses below. For tracking on terrain, the policy and the transition also receive a local height map \(P_t\), defined in Sec.~\ref{sec:dynamics}.
In this loop, policy improvement relies directly on the accuracy of $\hat s_{t+1}$ and its Jacobians, motivating the physics-informed transition model below.

%

\subsection{Learning a Physics-Informed World Model}
\label{sec:dynamics}
The world model $f_\phi$ acts as a differentiable surrogate of the simulator, supplying predicted rollouts and analytic Jacobians for policy optimization. Because the robot's kinematics, inertia, and PD actuation are known, $f_\phi$ avoids re-learning rigid-body mechanics from scratch: it learns only the unknown interaction forces and computes the resulting motion analytically (Fig.~\ref{fig:dynamimic-pipeline}a).
Specifically, the transition \(f_\phi\) in Eq.~\eqref{eq:method-io} consists of a learned force predictor \(\mathcal F_\phi\) and a differentiable rigid-body integrator \(I_{\mathrm{phys}}\):
\begin{equation}
\begin{aligned}
(W_t,\tau_{\mathrm{res},t}) &= \mathcal F_\phi(x_t),\\
\hat s_{t+1} &= I_{\mathrm{phys}}(s_t,a_t,\xi,W_t,\tau_{\mathrm{res},t}).
\end{aligned}
\label{eq:force-network}
\end{equation}
%
Below, we describe the four components that make this model physically grounded and trainable: (i) $\mathcal F_\phi$ extracts kinematic and terrain features $x_t$ to predict sub-step contact wrenches $W_t$ and a generalized residual $\tau_{\mathrm{res},t}$; (ii) a structured output layer constrains $W_t$ to unilateral normal loads and friction cones; (iii) $I_{\mathrm{phys}}$ maps these loads through link Jacobians to integrate the next state $\hat s_{t+1}$, from which the next observation, tracking reward, and soft termination follow analytically; and (iv) multi-step state and wrench losses jointly fit $\phi$ on replay windows.

\paragraph{Predicting contact wrenches.}
To resolve contact ambiguity—such as distinguishing a wrist hovering near the floor from one actively bearing weight—$\mathcal F_\phi$ must observe limb kinematics alongside the commanded action.
The state \(s_t=(Q_t,\nu_t)\) contains the base position, base orientation as a unit quaternion, and 29 joint angles in \(Q_t\). The generalized velocity \(\nu_t\in\mathbb R^{35}\) contains six base and 29 joint components. Let \(\mathcal C\) be the 15 candidate support links. On level ground, \(x_t\in\mathbb R^{238}\) comprises heading-invariant state features, the action, the episode parameters, and each candidate link's height, position relative to the base, and linear velocity. A wrist may be close to the floor both before contact and while bearing weight. Its velocity and the current action give the predictor additional evidence about the load at that wrist.

For terrain-conditioned tracking, we append a 64-dimensional convolutional encoding of the local height map \(P_t\) to these features. We sample the \(21\times21\) grid at 0.1\,m spacing in the base heading frame, with heights relative to the base. At each imagined step, bilinear interpolation at the predicted base pose updates the map for the next prediction.

Once per 20\,ms control step, the network predicts \(W_t\in\mathbb R^{4\times15\times6}\). Entry \(W_t[k,b]=\lambda_{b,t}^{(k)}=(F_{b,t}^{(k)},n_{b,t}^{(k)})\) is the force and moment about the origin of link \(b\in\mathcal C\) for the \(k\)-th 5\,ms interval, \(k\in\{0,1,2,3\}\). The four wrenches allow a link's load to change while the action is held fixed. The network also predicts one \(\tau_{\mathrm{res},t}\in\mathbb R^{35}\) for the entire control step to account for model mismatch.

\paragraph{Constraining wrench predictions.}
Unconstrained predictions may produce negative normal forces or
horizontal forces that exceed the friction limit. We therefore
transform the raw network outputs to enforce these constraints
and bound the contact moments. For each link \(b\in\mathcal C\)
and interval \(k\), let \(u_z\in\mathbb R\),
\(u_{xy}\in\mathbb R^2\), and \(u_n\in\mathbb R^3\) denote the
raw neural network outputs for vertical force, horizontal force, and moment,
respectively. For clarity, we suppress the link, time, and interval
indices on these outputs and the wrench components. We set
\begin{equation}
\begin{aligned}
F_z &= mg\,\log\!\left(1+\exp(u_z-2)\right),\\
F_{xy} &= \mu F_z\,\frac{\tanh\rho}{\rho}\,u_{xy},
\qquad \rho=\max(\|u_{xy}\|_2,10^{-6}),\\
n &= F_z\,(\ell_{b,x},\ell_{b,y},\mu\ell_{b,z})
              \odot\tanh(u_n).
\end{aligned}
\label{eq:contact-decoder}
\end{equation}
Here \(mg\) is the robot's weight, \(\mu\) is the friction coefficient,
and \(\ell_{b,x},\ell_{b,y},\ell_{b,z}\) are link-specific length scales
that bound the contact moments. The symbol \(\odot\) denotes elementwise
multiplication, and \(\tanh\) is applied elementwise to vectors.
This formulation ensures \(F_z\geq0\) and
\(\|F_{xy}\|_2\leq\mu F_z\), with moment bounds proportional to \(F_z\).
Thus, a lightly loaded link can transmit only a small horizontal
force and moment.
\paragraph{Predicting the next states.}
We combine the predicted contact wrenches and the network-predicted
residual generalized force with the robot dynamics to predict the
next state. Each control step lasts 20\,ms and uses two integration timesteps
of 10\,ms each, indexed by \(i\in\{0,1\}\).
Let \(\bar\lambda_{b,t}^{(i)}\) denote the average contact wrench
for link \(b\) at integration timestep \(i\), and \(J_{b,t}^{(i)}\)
the spatial Jacobian expressed at the same reference point and
in the same coordinate frame as the wrench.
The Jacobian maps generalized velocity to linear and angular velocity,
in the same order as the force and moment in the wrench.
The total applied generalized force is
\begin{equation}
\begin{aligned}
\tau_{\mathrm{tot},t}^{(i)}
&=\tau_{\mathrm{act},t}^{(i)}+\tau_{\mathrm{pass},t}^{(i)}
\\[-2pt]
&\quad+\sum_{b\in\mathcal C}(J_{b,t}^{(i)})^\top
\bar\lambda_{b,t}^{(i)}+\tau_{\mathrm{res},t}.
\end{aligned}
\label{eq:applied-load}
\end{equation}
The PD controller produces the actuator contribution
\(\tau_{\mathrm{act},t}^{(i)}\) from the action's joint targets.
The passive contribution \(\tau_{\mathrm{pass},t}^{(i)}\)
includes joint damping. The Jacobian transpose maps each link
wrench to a generalized force acting on the base and joints.

We use semi-implicit Euler integration to update the robot's velocity
and configuration, with implicit treatment of damping and
velocity-dependent actuator torques. Within each control step,
we omit the time index \(t\) on the configuration, velocity, and
dynamics matrices. Starting from
\((Q^{(0)},\nu^{(0)})=(Q_t,\nu_t)\), we perform two updates,
indexed by \(i\in\{0,1\}\), with timestep \(\Delta=10\,\mathrm{ms}\):
\begin{equation}
\begin{aligned}
(M^{(i)}+\Delta B^{(i)})(\nu^{(i+1)}-\nu^{(i)})
   &=\Delta\bigl(\tau_{\mathrm{tot},t}^{(i)}-h^{(i)}\bigr),\\
Q^{(i+1)}&=Q^{(i)}\oplus(\Delta\nu^{(i+1)}).
\end{aligned}
\label{eq:configuration-step}
\end{equation}
Here \(M\) is the generalized mass matrix, \(h\) contains gravity
and velocity-dependent terms, and \(B\) accounts for passive damping
and the velocity dependence of unsaturated actuator torques in the
implicit solve. Superscript \((i)\) denotes quantities evaluated
at the start of integration timestep \(i\), while \((i+1)\)
denotes the updated state. The operator \(\oplus\) advances base
translation and joint angles and composes the base orientation
with its angular increment. After the two updates, the world-model
prediction is \(\hat s_{t+1}=(Q^{(2)},\nu^{(2)})\).

\paragraph{Fitting the world model.}
During simulator rollouts, we store states, actions, and per-link
contact wrenches in a replay buffer for world-model training.
We sample windows of eight consecutive states, together with
the corresponding actions and contact wrenches, from this buffer.
Starting from the first state of each window, we roll the model
forward for seven steps under the recorded actions and supervise
the predicted states and per-link contact wrenches
(Fig.~\ref{fig:dynamimic-pipeline}b). Trajectory supervision penalizes
multi-step drift, while wrench supervision resolves how forces
are distributed across contacts, such as between the two feet
during double support. The training objective combines trajectory
supervision, link-level wrench supervision, and residual regularization:
%
%
\begin{equation}
\mathcal L_{\mathrm{WM}}
=\mathcal L_{\mathrm{state}}+\mathcal L_{\mathrm{wrench}}
+10^{-3}\mathcal L_{\mathrm{res}}.
\label{eq:model-loss}
\end{equation}
The state term averages scaled squared errors in base position, base orientation, joint angles, and joint velocities over the predicted states. Each error channel is divided by its group's nominal error before squaring. For orientation, \(\hat R\) and \(R^\star\) are the predicted and simulator base rotation matrices. Their squared discrepancy is \(\tfrac12\|\hat R-R^\star\|_F^2\), where \(\|\cdot\|_F\) denotes the Frobenius norm. For wrench supervision, we compare forces in world coordinates and moments about their respective link origins. A simulator force \(F^\star\) applied at \(c\), with moment \(n_c^\star\) about \(c\), gives the target moment \(n_b^\star=n_c^\star+(c-p_b)\times F^\star\) about the simulator link origin \(p_b\). The wrench loss is
\begin{equation}
\mathcal L_{\mathrm{wrench}}
=\left\langle
\frac{\|\hat F-F^\star\|_2^2}{(mg)^2}
+
\frac{\|\hat n_b-n_b^\star\|_2^2}{(mg\ell_0)^2}
\right\rangle,\qquad \ell_0=0.1\,\mathrm m.
\label{eq:link-wrench-loss}
\end{equation}
The brackets \(\langle\cdot\rangle\) average over training windows, predicted transitions, wrench samples, and candidate links. Links out of contact have zero target wrench. For \(\mathcal L_{\mathrm{res}}\), we divide each of the 35 components of \(\tau_{\mathrm{res},t}\) by its numerical scale of 5 in the generalized-force coordinates, square the result, and average over components, predicted transitions, and training windows. 


\subsection{Learning the Tracking Policy}
\label{sec:training}
We now use the world model learned in section \ref{sec:dynamics} to improve the tracking policy. Starting from a simulator state sampled from replay, the policy chooses its own actions and the frozen model predicts their consequences for 32 control steps. We optimize the tracking rewards over these steps and use a critic to estimate the remaining return from the final predicted state. During the actor update, gradients flow through the predicted contact wrenches and rigid-body dynamics to the actions that produced the motion, as in Fig.~\ref{fig:dynamimic-pipeline}c.
\paragraph{Rolling out the policy.}
For the policy update (Fig.~\ref{fig:dynamimic-pipeline}c), we initialize each imagined rollout from a replay state and its observation history. A reference encoder embeds a window of ten reference frames, and an action decoder combines that embedding with the ten most recent proprioceptive observations. Each observation contains projected gravity, base angular velocity, joint positions and velocities, and the previous action. For terrain-conditioned tracking, the actor and critic each encode the current local height map. At every imagined step, the policy selects an action, DynaMimic predicts the next state, and the resulting observation enters the history for the next decision.

\paragraph{Updating the actor.}
%
To improve tracking performance over these imagined trajectories, we update $\theta$ by backpropagating the critic-bootstrapped tracking return through the frozen world model and critic. At rollout step $j$, let $\hat r_{j+1}$ denote the tracking reward (measuring base and body tracking while penalizing action smoothness and joint-limit violations), $d_{j+1}$ the soft termination probability from base-height and tilt errors, and $c_{j+1}=c_j(1-d_{j+1})$ with $c_0=1$ the cumulative survival weight. With discount $\gamma=0.99$ and critic prediction $V_j=V_\psi(\hat s_j,g_j,\xi)$ (augmented with the height map at $\hat s_j$ on terrain), we cap the terminal bootstrap at the maximum attainable return $V_{\max}=r_{\max}/(1-\gamma)$ via $\bar V_H=\min(V_H,V_{\max})$ and maximize
\begin{equation}
 J(\theta) = \mathbb E\left[\sum_{j=0}^{H-1}\gamma^j c_j \hat r_{j+1} + \gamma^H c_H \bar V_H\right],
 \label{eq:actor-loss}
\end{equation}
averaged over replay-seeded rollouts. Holding $\phi$ and $\psi$ fixed, backpropagation-through-time passes gradients through the rigid-body updates and predicted wrenches back to every action along the trajectory.

\paragraph{Fitting the critic.}
%
To provide an accurate terminal value $\bar V_H$ for the actor objective in Eq.~\eqref{eq:actor-loss}, we fit $V_\psi$ on the same imagined transitions using multi-step $\lambda$-returns ($\lambda=0.95$) that balance bias and horizon length. Starting from the uncapped terminal value $G_H^\lambda = V_H$, we compute targets backward for $j=H-1,\dots,0$:
\begin{equation}
 G_j^\lambda = \hat r_{j+1} + \gamma(1-d_{j+1})\bigl[(1-\lambda)V_{j+1} + \lambda G_{j+1}^\lambda\bigr],
 \label{eq:critic-target}
\end{equation}
cap each $G_j^\lambda$ at $V_{\max}$, and minimize the mean squared error between $V_j$ and the detached target. At deployment, both $f_\phi$ and $V_\psi$ are discarded: only $\pi_\theta$ runs on the robot, mapping reference motions, sensor history, and optional local terrain maps directly to joint commands.
\vspace{-1mm}
\section{Experiments}
\label{sec:experiments}

\begin{table*}[!t]
\centering
\small
\setlength{\tabcolsep}{2pt}
\begin{tabular*}{\linewidth}{@{\extracolsep{\fill}}llrrrrrr@{}}
\toprule
Evaluation set & Method & Ret. $\uparrow$ & \shortstack{MPJPE $\downarrow$} & \shortstack{Root-relative\\MPJPE $\downarrow$} & Vel. $\downarrow$ & Acc. $\downarrow$ & WM $\downarrow$ \\
\midrule
LAFAN1 (60 full motions) & SuperMimic & 0.8256 & 1,284.42 & 41.52 & 5.189 & 1.903 & 0.007909 \\
 & DynaMimic & \textbf{0.8301} & \textbf{1,045.39} & \textbf{39.62} & \textbf{5.022} & \textbf{1.881} & \textbf{0.006790} \\
\addlinespace[2pt]
LAFAN1 (18 challenging intervals) & SuperMimic & 0.7815 & 606.92 & 61.53 & 6.831 & 2.406 & 0.015870 \\
 & DynaMimic & \textbf{0.8123} & \textbf{513.07} & \textbf{49.41} & \textbf{6.242} & \textbf{2.293} & \textbf{0.009016} \\
\bottomrule
\end{tabular*}
\caption{\textbf{Motion tracking on LAFAN1.} }
\label{tab:tracking-50k}
\end{table*}

\begin{figure}[!t]
\centering
\begin{minipage}[t]{0.49\columnwidth}
  \centering
  \includegraphics[width=\linewidth]{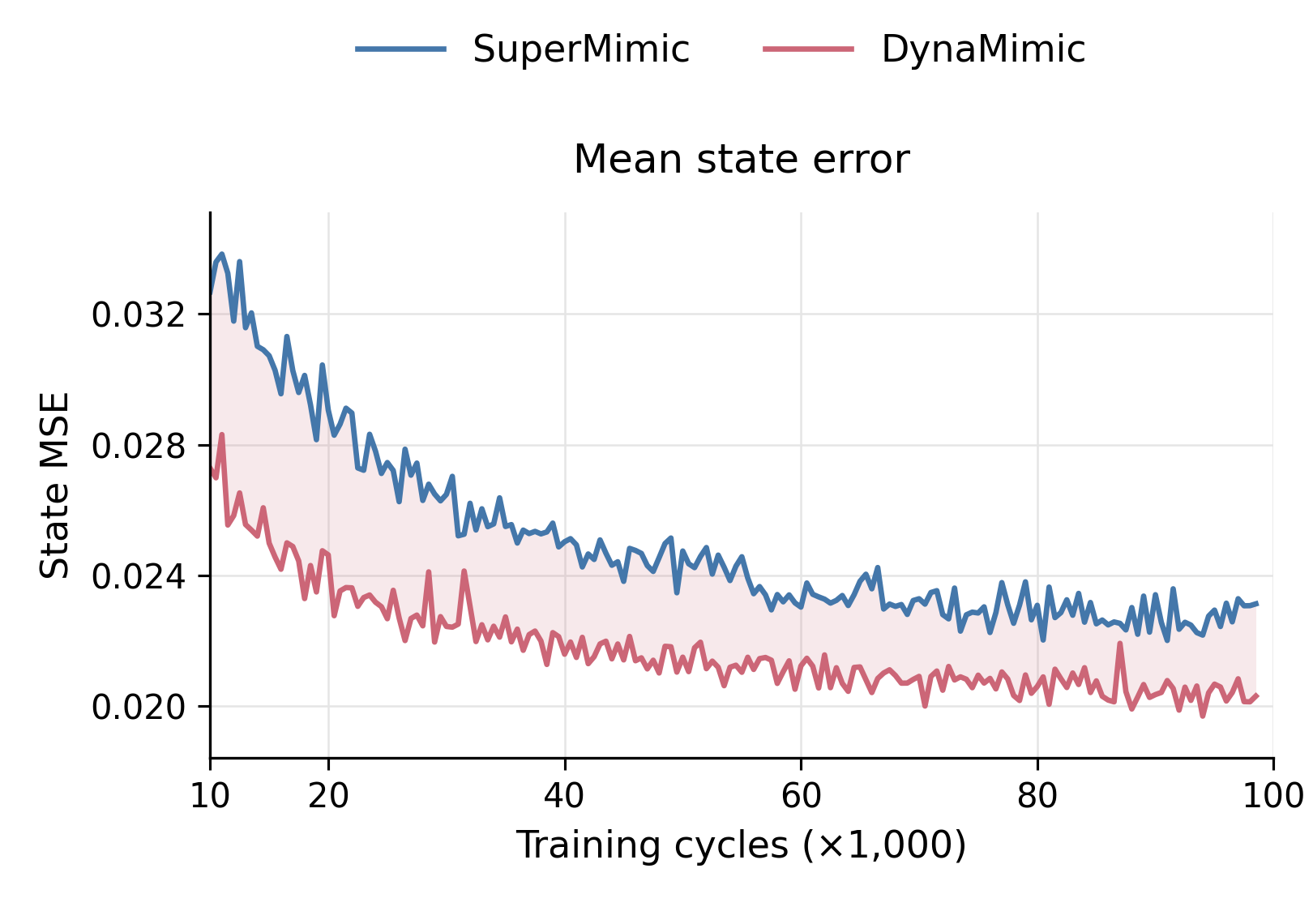}\\[-2pt]
  {\small (a)}
\end{minipage}\hfill
\begin{minipage}[t]{0.49\columnwidth}
  \centering
  \includegraphics[width=\linewidth]{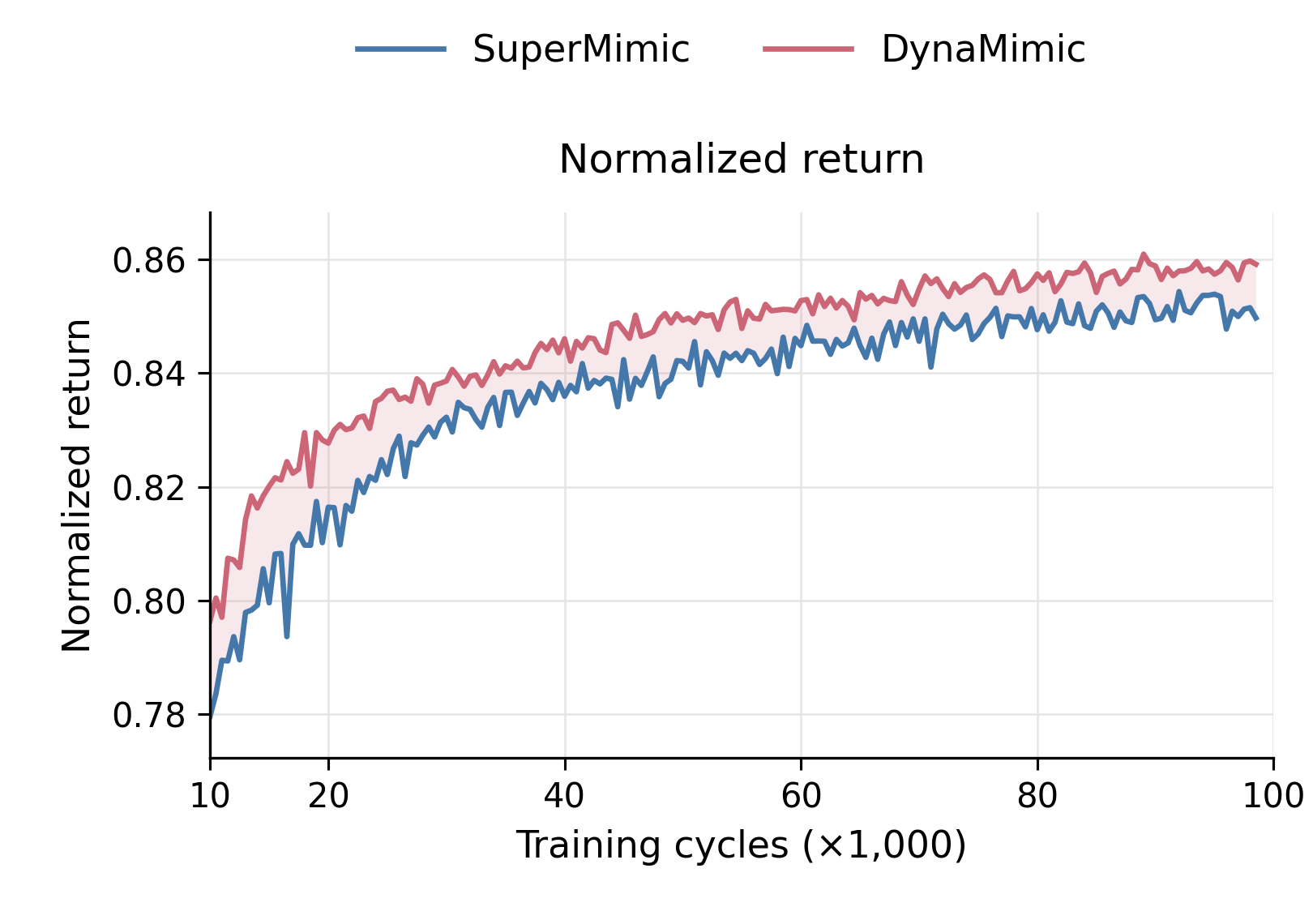}\\[-2pt]
  {\small (b)}
\end{minipage}
\caption{\textbf{LAFAN1 training curves.} (a) World model MSE. (b) Normalized return. Both curves span 10,000 to 100,000.}
\label{fig:lafan-training}
\end{figure}


\begin{table}[!tb]
\centering
\footnotesize
\setlength{\tabcolsep}{2pt}
\resizebox{\columnwidth}{!}{\begin{tabular}{@{}lrrrrr@{}}
\toprule
Method & Ret. $\uparrow$ & MPJPE $\downarrow$ & Vel. $\downarrow$ & Acc. $\downarrow$ & WM $\downarrow$ \\
\midrule
\multicolumn{6}{@{}l}{\textit{No detected non-foot contact, lower intensity}} \\
SuperMimic & 0.9217 & 172.77 & 3.349 & \textbf{1.057} & 0.003026 \\
DynaMimic & \textbf{0.9241} & \textbf{160.85} & \textbf{3.287} & 1.061 & \textbf{0.002783} \\
\textit{Gain} & $+0.0023$ & $+11.91$ & $+0.062$ & $-0.004$ & $+0.000243$ \\
\addlinespace[2pt]
\multicolumn{6}{@{}l}{\textit{No detected non-foot contact, higher intensity}} \\
SuperMimic & 0.8414 & 315.30 & 6.889 & 2.660 & 0.005805 \\
DynaMimic & \textbf{0.8449} & \textbf{292.61} & \textbf{6.669} & \textbf{2.634} & \textbf{0.005108} \\
\textit{Gain} & $+0.0034$ & $+22.69$ & $+0.220$ & $+0.025$ & $+0.000697$ \\
\addlinespace[2pt]
\multicolumn{6}{@{}l}{\textit{Detected non-foot contact, lower intensity}} \\
SuperMimic & 0.8424 & \textbf{177.50} & 4.897 & 1.711 & 0.022434 \\
DynaMimic & \textbf{0.8495} & 183.26 & \textbf{4.778} & \textbf{1.684} & \textbf{0.020694} \\
\textit{Gain} & $+0.0072$ & $-5.77$ & $+0.119$ & $+0.027$ & $+0.001740$ \\
\addlinespace[2pt]
\multicolumn{6}{@{}l}{\textit{Detected non-foot contact, higher intensity }} \\
SuperMimic & 0.7882 & 272.38 & 7.503 & 2.856 & 0.025922 \\
DynaMimic & \textbf{0.8008} & \textbf{252.65} & \textbf{7.194} & \textbf{2.782} & \textbf{0.020022} \\
\textit{Gain} & $+0.0126$ & $+19.73$ & $+0.309$ & $+0.074$ & $+0.005899$ \\
\bottomrule
\end{tabular}}
\caption{\textbf{LAFAN1 tracking by contact and intensity.}}
\label{tab:lafan-contact-gains}
\end{table}

\begin{figure*}[t]
\centering
\begingroup
\setlength{\tabcolsep}{1.8pt}
\setlength{\fboxsep}{0pt}
\setlength{\fboxrule}{1pt}
\definecolor{lafanFail}{rgb}{0.64,0.23,0.20}
\definecolor{lafanEmpty}{rgb}{0.945,0.953,0.961}
\def\lafanframewidth{\dimexpr0.179\textwidth-2\tabcolsep\relax}
\def\lafanshot#1#2{\raisebox{-0.5\height}{\includegraphics[page=#1,trim=#2,clip,width=\lafanframewidth]{figures_lafan_failure/fig3_sage_oak.pdf}}}
\def\lafanlabel#1{\parbox[c]{0.105\textwidth}{\raggedright\small\textbf{#1}}}
\small
\begin{tabular}{@{}lccccc@{}}
\multicolumn{6}{@{}l}{\textbf{(a) LAFAN1 ground-contact recovery}} \\[1pt]
& 0.00\,s & 0.72\,s & 1.72\,s & 2.22\,s & 10.00\,s \\[1pt]
\lafanlabel{Reference}
& \lafanshot{1}{18bp 24.5bp 236.5bp 117bp}
& \lafanshot{2}{26bp 32bp 231bp 111.5bp}
& \lafanshot{3}{17bp 49bp 237.5bp 92.5bp}
& \lafanshot{4}{24bp 62.5bp 223bp 73bp}
& \lafanshot{5}{199.5bp 66.5bp 0bp 31bp} \\[1pt]
\lafanlabel{SuperMimic}
& \lafanshot{6}{18bp 24.5bp 236.5bp 117bp}
& \lafanshot{7}{26bp 32bp 231bp 111.5bp}
& \lafanshot{8}{17bp 49bp 237.5bp 92.5bp}
& \makebox[\lafanframewidth][c]{\raisebox{-0.5\height}{\fcolorbox{lafanFail}{white}{\includegraphics[page=9,trim=24bp 62.5bp 223bp 73bp,clip,width=\lafanframewidth]{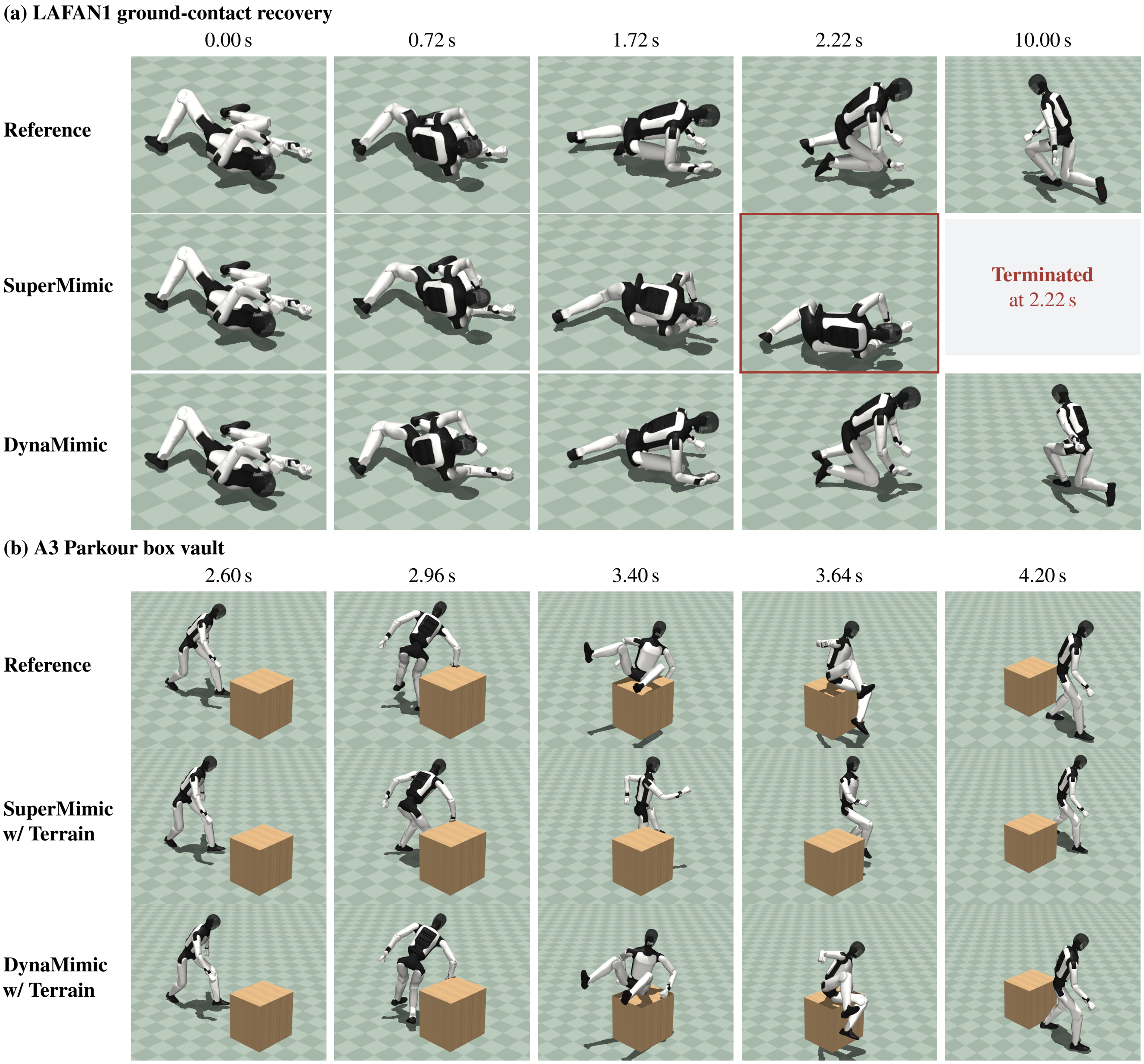}}}}
& \colorbox{lafanEmpty}{\parbox[c][0.12\textwidth][c]{\lafanframewidth}{\centering\color{lafanFail}\small\textbf{Terminated}\\at 2.22\,s}} \\[1pt]
\lafanlabel{DynaMimic}
& \lafanshot{10}{18bp 24.5bp 236.5bp 117bp}
& \lafanshot{11}{26bp 32bp 231bp 111.5bp}
& \lafanshot{12}{17bp 49bp 237.5bp 92.5bp}
& \lafanshot{13}{24bp 62.5bp 223bp 73bp}
& \lafanshot{14}{199.5bp 66.5bp 0bp 31bp} \\
\noalign{\vskip 3pt}
\multicolumn{6}{@{}l}{\textbf{(b) A3 Parkour box vault}} \\[1pt]
& 2.60\,s & 2.96\,s & 3.40\,s & 3.64\,s & 4.20\,s \\[1pt]
\lafanlabel{Reference} & \lafanshot{15}{51bp 42.5bp 96bp 13bp} & \lafanshot{16}{78bp 42.5bp 84bp 25bp} & \lafanshot{17}{84bp 42.5bp 55.5bp 7bp} & \lafanshot{18}{99bp 42.5bp 33bp 1bp} & \lafanshot{19}{92bp 0.5bp 0bp 11bp} \\[1pt]
\lafanlabel{SuperMimic\\w/ Terrain} & \lafanshot{20}{51bp 42.5bp 96bp 13bp} & \lafanshot{21}{78bp 42.5bp 84bp 25bp} & \lafanshot{22}{84bp 42.5bp 55.5bp 7bp} & \lafanshot{23}{99bp 42.5bp 33bp 1bp} & \lafanshot{24}{92bp 0.5bp 0bp 11bp} \\[1pt]
\lafanlabel{DynaMimic\\w/ Terrain} & \lafanshot{25}{51bp 42.5bp 96bp 13bp} & \lafanshot{26}{78bp 42.5bp 84bp 25bp} & \lafanshot{27}{84bp 42.5bp 55.5bp 7bp} & \lafanshot{28}{99bp 42.5bp 33bp 1bp} & \lafanshot{29}{92bp 0.5bp 0bp 11bp} \\
\end{tabular}
\endgroup
\caption{\textbf{Floor recovery and box vault.}
\textbf{(a) LAFAN1 floor recovery.} DynaMimic tracks the reference as the humanoid rolls onto its side, pushes up from the floor, and rises to one knee. It completes the 10\,s segment, whereas SuperMimic remains on the floor and its rollout ends at 2.22\,s.
\textbf{(b) A3 Parkour box vault.}  DynaMimic reproduces the hand placement and leg swing. SuperMimic with terrain input keeps its torso more upright and misses the sideways leg swing.}
\label{fig:lafan-push-recovery}
\end{figure*}

\begin{table}[!tb]
\centering
\small
\setlength{\tabcolsep}{2pt}
\resizebox{\columnwidth}{!}{%
\begin{tabular}{@{}lrrrrr@{}}
\toprule
Method & Ret. $\uparrow$ & MPJPE $\downarrow$ & Vel. $\downarrow$ & Acc. $\downarrow$ & WM $\downarrow$ \\
\midrule
SuperMimic & 0.6734 & 363.24 & 9.78 & 2.95 & 0.03463 \\
SuperMimic w/ Terrain & 0.9068 & 66.15 & 4.94 & 2.01 & 0.00793 \\
DynaMimic w/ Terrain& \textbf{0.9186} & \textbf{56.28} & \textbf{4.55} & \textbf{1.91} & \textbf{0.00454} \\
\bottomrule
\end{tabular}}
\caption{\textbf{Terrain-aware motion tracking on A3 Parkour.}}
\label{tab:a3-parkour-50k}
\end{table}

\begin{table}[!tb]
\centering
\small
\setlength{\tabcolsep}{2pt}
\resizebox{\columnwidth}{!}{%
\begin{tabular}{@{}lrrrrr@{}}
\toprule
Method & Ret. $\uparrow$ & MPJPE $\downarrow$ & Vel. $\downarrow$ & Acc. $\downarrow$ & WM $\downarrow$ \\
\midrule
\multicolumn{6}{@{}l}{\textit{Overall (9,736 clips)}} \\
SuperMimic & 0.4598 & 273.19 & 9.1115 & 2.9884 & 0.07155 \\
SuperMimic w/ Terrain & 0.8477 & 141.88 & 4.6099 & 1.6174 & 0.01125 \\
DynaMimic w/ Terrain & \textbf{0.8604} & \textbf{126.57} & \textbf{4.6027} & \textbf{1.6167} & \textbf{0.01082} \\
\bottomrule
\end{tabular}}
\caption{\textbf{Terrain-aware motion tracking on GRAIL.}}
\label{tab:grail-terrain-tracking}
\end{table}

\begin{table}[t]
\centering
\footnotesize
\setlength{\tabcolsep}{2pt}
\resizebox{\linewidth}{!}{%
\begin{tabular}{@{}lrrrrr@{}}
\toprule
Model & Ret. $\uparrow$ & MPJPE $\downarrow$ & Vel. $\downarrow$ & Acc. $\downarrow$ & WM $\downarrow$ \\
\midrule
SuperMimic w/ link inputs & 0.8617 & 258.52 & 5.475 & 1.973 & 0.007488 \\
DynaMimic w/o link inputs & 0.8628 & 243.82 & 5.449 & 1.993 & 0.009705 \\
DynaMimic w/o forward dynamics & 0.8513 & 277.50 & 5.647 & 2.008 & 0.011705 \\
DynaMimic w/o contact loss & 0.8673 & 248.64 & 5.384 & 1.975 & 0.007437 \\
DynaMimic & \textbf{0.8686} & \textbf{240.63} & \textbf{5.345} & \textbf{1.955} & \textbf{0.007368} \\
\bottomrule\end{tabular}}
\caption{\textbf{Ablation study on LAFAN1.}}
\label{tab:lafan-world-model-ablation}
\end{table}

\paragraph{Experimental setup.}
We train one policy per method and dataset on a single NVIDIA H800 GPU. 

\subsection{Evaluation Metrics}
\label{sec:eval-metrics}

\paragraph{Tracking error.}
Let $p_{tb}$ and $p^{\mathrm{ref}}_{tb}$ denote the simulated and reference positions (m) of link $b$ at step $t$ in world coordinates. Mean per-joint position error (MPJPE, mm) averages over all $T$ executed steps and $B=14$ tracked links:
\begin{equation}
  E_p = \frac{1000}{TB}\sum_{t=1}^{T}\sum_{b=1}^{B}
  \lVert p_{tb}-p^{\mathrm{ref}}_{tb}\rVert_2 .
\end{equation}
To compute root-relative MPJPE, we subtract the pelvis position from each tracked link position in each pose before comparing the simulated and reference poses at every time step. For velocity error (Vel., mm/frame), we replace positions by $\Delta t\,v_{tb}$ and its reference, where $v_{tb}$ is world-frame link velocity and $\Delta t=0.02$\,s. Acceleration error (Acc., mm/frame$^2$) compares successive differences of these scaled velocities over the $T-1$ frame pairs. Each rollout contributes equally to the reported means.
\vspace{-1mm}
\paragraph{Return and completion.}
Normalized return (Ret.) is the dimensionless ratio $R=\sum_{t=1}^{T}r_t/(N r_{\max})$, where $r_t$ is the reward, $r_{\max}$ its per-step maximum, and $N$ the reference length in steps. Unexecuted steps earn zero reward. 

\vspace{-1mm}
\paragraph{World model prediction.}
WM MSE is the mean squared error of the normalized state, averaged over seven open-loop prediction steps. For each checkpoint, its policy generates a simulator trajectory. Starting from a state on that trajectory, the corresponding world model replays the seven recorded actions and predicts the seven subsequent simulator states. Thus WM MSE measures prediction accuracy on the trajectories each policy visits; the action sequences can differ between methods. In the tables, $\uparrow$ indicates that higher values are better and $\downarrow$ that lower values are better.

\vspace{-2mm}
\subsection{Datasets and Evaluation Protocol}
\label{sec:eval-settings}

\paragraph{LAFAN1.}
LAFAN1~\cite{harvey2020robust} contains 77 motion-capture sequences from five performers, covering locomotion, dance, falls, and recoveries. We exclude motions that rely on obstacles because they are difficult to reproduce on level ground without the corresponding scene geometry. The remaining 60 sequences last 102--306 seconds each. These include the 18 challenging intervals identified by OmniXtreme~\cite{wang2026omnixtreme}.

\vspace{-1mm}

\paragraph{A3 Parkour.}
We collected A3 Parkour, a dataset of 22 parkour motions paired with the geometry of the static box used in each recording. The boxes vary in height and length. The motions include hand-supported vaults, climbs onto the box, and dismounts that end in a roll.

\vspace{-1mm}

\paragraph{GRAIL.}\label{sec:terrain-evaluation}
GRAIL~\cite{xie2026grailgeneratinghumanoidlocomanipulation} provides over 34,000 simulated humanoid motion sequences in its public release, spanning object pickup, sitting, and terrain traversal. We use the three terrain categories, which pair each motion with the geometry of the curb, slope, or staircase traversed by the humanoid.

\paragraph{Evaluation.}
We train a separate tracking policy for each dataset and share it across reference motions within that dataset. On A3 Parkour and GRAIL, we evaluate the reference libraries used for training (22 motions and 9,736 clips, respectively). The 18 challenging LAFAN1 intervals are drawn from the 60 full motions. For each full motion, we initialize the robot at the first reference pose and velocity and track continuously on level ground (LAFAN1) or with the paired scene geometry (A3 Parkour and GRAIL). For each challenging LAFAN1 interval, we reset to its first reference pose and velocity, so tracking starts without drift from the preceding motion. Each rollout stops at the end of the sequence or interval, or earlier if the robot triggers a failure.

\vspace{-1mm}

\paragraph{LAFAN1 motion groups.}
We divide the LAFAN1 motion sequences into 1,307 segments of up to 10 seconds and start each rollout from the segment’s first reference frame. This keeps tracking failures during a brief, difficult motion from carrying over into later segments of the same sequence. We group segments by whether they exhibit potential non-foot ground contact and use the median motion intensity to distinguish lower- and higher-intensity segments.


\vspace{-1mm}

\subsection{Results}
\label{sec:lafan-controls}

\paragraph{Level-ground tracking.}
DynaMimic and SuperMimic both complete all 60 full motions from LAFAN1~\cite{harvey2020robust} and all 18 challenging intervals identified by OmniXtreme~\cite{wang2026omnixtreme} (Table~\ref{tab:tracking-50k}). DynaMimic achieves higher return and lower world-frame MPJPE on both sets. It also tracks body pose relative to the pelvis more accurately, with a larger reduction in root-relative MPJPE on the challenging intervals. At most training checkpoints, DynaMimic achieves higher return and lower WM error.(Fig.~\ref{fig:lafan-training}).

\vspace{-1mm}

\paragraph{Contact analysis.}
We group LAFAN1 segments by motion intensity and whether the reference motion involves non-foot ground contact (Table~\ref{tab:lafan-contact-gains}). DynaMimic achieves higher return and lower WM error than SuperMimic in every group. MPJPE also decreases in every group except the low-intensity group with non-foot contact. DynaMimic shows its clearest advantage on high-intensity motions with non-foot contact, where it achieves the largest gain in return and the largest reductions in velocity, acceleration, and WM errors. Figure~\ref{fig:lafan-push-recovery}(a) illustrates a floor recovery: from the same initial state, DynaMimic completes the motion while SuperMimic terminates early.

\vspace{-1mm}

\paragraph{Terrain-conditioned tracking.}
Adding a local height map to SuperMimic substantially reduces MPJPE on both A3 Parkour and GRAIL~\cite{xie2026grailgeneratinghumanoidlocomanipulation} (Tables~\ref{tab:a3-parkour-50k} and~\ref{tab:grail-terrain-tracking}). With the same terrain input, DynaMimic further improves return and reduces MPJPE and WM error on both datasets. On A3 Parkour, both terrain-aware policies complete all 22 motions, but DynaMimic tracks position, velocity, and acceleration more accurately. On GRAIL, DynaMimic also completes a larger fraction of clips.

\vspace{-1mm}

\paragraph{Ablation study.}
We evaluate the contributions of link inputs, forward dynamics, and contact wrench supervision on the same ten-second LAFAN1 segments (Table~\ref{tab:lafan-world-model-ablation}). The full DynaMimic model achieves higher return and lower tracking and WM errors than SuperMimic with the same link inputs. Removing either the link inputs or contact wrench supervision reduces tracking accuracy. Removing forward dynamics causes the largest drop in return and the largest increase in every error metric, indicating that it contributes most to tracking performance among the components tested.

\vspace{-1mm}

\paragraph{Real-world deployment.}
On the physical A3 humanoid, we run the trained tracking policy without the world model. The robot rises from the floor, performs a cartwheel, and vaults over a box (Fig.~\ref{fig:world-model}). These motions demonstrate changes of support on ground and an elevated obstacle.

\vspace{-1mm}
\section{Conclusion}
DynaMimic trains humanoid motion tracking policies with a differentiable world model that learns contact wrenches and a residual generalized force, then uses the robot's rigid-body dynamics to predict the resulting motion. DynaMimic reduces MPJPE by 18.6\% relative to SuperMimic on LAFAN1. When both methods receive the same local height maps, DynaMimic reduces this error by 14.9\% on A3 Parkour and 10.8\% on GRAIL terrain clips. Our trained policies can run directly on a physical A3 humanoid, enabling it to rise from the floor, perform a cartwheel, and vault over a box.

\clearpage
\section*{LLM Usage Statement}
We used ChatGPT for manuscript polishing and assistance with writing and debugging code.
{\small

\bibliographystyle{ieeenat_fullname}
\bibliography{main}
}
\end{document}